\documentclass[runningheads,a4paper]{llncs}

\pdfoutput=1
\usepackage{amssymb}
\usepackage{graphicx}
\usepackage{amsmath}
\usepackage{textcomp}
\usepackage{hhline}

\usepackage{url}
\urldef{\mailsa}\path{jina1407@gmail.com}
\newcommand{\keywords}[1]{\par\addvspace\baselineskip
\noindent\keywordname\enspace\ignorespaces#1}

\begin{document}

\setlength{\parskip}{0pt}

\mainmatter  

\title{A Visual Measure of Changes to Weighted Self-Organizing Map Patterns}

\titlerunning{A Visual Measure of Changes to Weighted SOM Patterns}

%
%
\author{Younjin Chung \and Joachim Gudmundsson \and Masahiro Takatsuka}
%
\authorrunning{Younjin Chung \and Joachim Gudmundsson \and Masahiro Takatsuka}

\institute{School of IT, Faculty of Engineering and IT\\
The University of Sydney, NSW 2006 Australia}

%
%

\toctitle{A Visual Measure of Changes to Weighted SOM Patterns}
\tocauthor{Younjin Chung, Joachim Gudmundsson, Masahiro Takatsuka}
\maketitle

\begin{abstract}
Estimating output changes by input changes is the main task in causal analysis. In previous work, input and output Self-Organizing Maps (SOMs) were associated for causal analysis of multivariate and nonlinear data. Based on the association, a weight distribution of the output conditional on a given input was obtained over the output map space. Such a weighted SOM pattern of the output changes when the input changes. In order to analyze the change, it is important to measure the difference of the patterns. Many methods have been proposed for the dissimilarity measure of patterns. However, it remains a major challenge when attempting to measure how the patterns change. In this paper, we propose a visualization approach that simplifies the comparison of the difference in terms of the pattern property. Using this approach, the change can be analyzed by integrating colors and star glyph shapes representing the property dissimilarity. Ecological data is used to demonstrate the usefulness of our approach and the experimental results show that our approach provides the change information effectively.

\end{abstract}

\section{Introduction} \label{sec1}

Analyzing causality is one of the central tasks of science since it influences decision making in such diverse domains as natural, social and health sciences. Causality is the relationship between two events, if changes of one (cause) trigger changes of the other (effect) \cite{may1970}. In our previous work \cite{chung2015}, a causal analysis model was developed for analyzing causality of multivariate and nonlinear data (unlabeled in nature). In that model, different Self-Organizing Maps (SOMs) \cite{Kohonen2001} for input and output data sets were networked using a weight association based on the connection prototype feature vector similarity. Given such SOMs, the similarity weights conditional on a given input could be assigned to the neurons of the output SOM. Such a weighted SOM pattern of the output is described by two information types: $(1)$ the weight distribution and $(2)$ the property (prototype feature vector) distribution. For assessing output changes by input changes, it is crucial to measure the property difference of weighted SOM patterns. 

There have been many attempts to measure the dissimilarity between two distributions (patterns) such as the Minkowski and the Shanon's entropy families \cite{cha2007,johnson2000,kullback2012}, the Quadratic Form Distance \cite{niblack1993} and the Earth Mover's Distance (EMD) \cite{rubner2000}. For weights with adaptive neurons in a weighted SOM pattern, we found the EMD to be the most suitable method in measuring the dissimilarity of weighted SOM patterns. The EMD, as the others, aims to provide a numerical value to define only a notion of the overall resemblance of patterns. It cannot differentiate between weighted SOM patterns if they have the same dissimilarity but different properties. Therefore, we introduce a method, called Property EMD (PEMD), to measure the property difference by individual feature differences in the pattern change. However, it is still difficult to represent and to compare the overall property dissimilarity in the change for high dimensional data. It is also difficult to observe possible feature values that gain the pattern change.   

Due to the limitations of quantitative approaches, we propose a visualization approach for measuring the property change of weighted SOM patterns along with the PEMD. Our visualization integrates colors and graphical shape objects of star glyph to represent the pattern information in the change. Using this approach, the property dissimilarity of weighted SOM patterns can be captured with the size and the direction of the change. Possible feature values that gain the pattern change can also be observed by exploring regions of interest in weighted SOM patterns. Ecological data is used to demonstrate that our approach is useful for the pattern property comparison in the pattern change. The experimental results show that our approach provides the change information considered for causal analysis in an effective visual way.

\section{Background} \label{sec2}

\begin{figure}[tb]
\begin{center}
\includegraphics[width=0.99\linewidth]{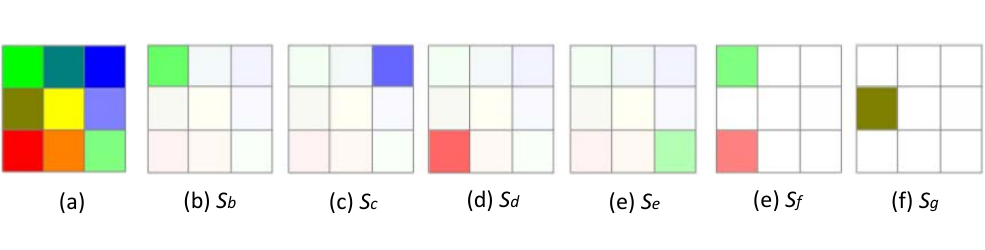} 
\caption{$(a)$ A $3\times3$ SOM of a RGB color space. $(b)$ - $(g)$: Six weighted SOM patterns of the SOM in $(a)$. The changes are made from $S_b$ to $S_c$ - $S_e$, and from $S_f$ to $S_g$.\label{fig1}}
\end{center}
\end{figure}

A Self-Organizing Map (SOM) \cite{Kohonen2001} projects high dimensional data onto a low dimensional (typically $2$-dimensional) grid map space. A set of neurons of the map, which are prototype feature vectors adaptively projected for original feature vectors, reflects the data properties. Using the causal analysis model in our previous work \cite{chung2015}, a weight distribution is estimated on the property distribution of the output SOM for a given input. Fig. \ref{fig1} shows a simple example illustrating two information types in a weighted SOM pattern: $(1)$ the weight distribution and $(2)$ the property distribution. The SOM in Fig. \ref{fig1}$(a)$ is used as it is easy for visualizing $3$-dimensional RGB color property and position. Based on the SOM, several weighted SOM patterns ($(b)$ - $(g)$ in Fig. \ref{fig1}) are created by different color opacity values (weights). Such a weighted SOM pattern can be depicted as $S = \{(v_i, w_i)\}$, $i = 1$,..,$n$, where $n$ is the number of neurons ($n=9$); $v_i$ is $i$th prototype feature vector ($v_i = \{(c_k)_i\}$, $k = 1$,..,$m$, where $m$ is the number of features ($m=3$); $c_k$ is the $k$th component of the prototype feature vector) and $w_i$ is $i$th weight, representing the two information types. 

The perceptual dissimilarity between two weighted SOM patterns in Fig. \ref{fig1} can be measured by observing the color properties in the highly weighted neurons. The patterns $S_c$ and $S_d$ are different patterns by their color properties although they have the same distance in relation to their changes from $S_b$. The pattern $S_g$ shows the change in two perspectives by the two different color properties highlighted in $S_f$. Such differences can be measured using the $3$-dimensional color property. Nonetheless, it is still difficult to estimate the size and the direction of the change with respect to the pattern property. Furthermore, it becomes harder to measure such differences in higher dimensions. 

There have been many methods to quantitatively measure the dissimilarity between two distributions (patterns) of high dimensional data. The most used families of functions are the Minkowski and the Shanon's entropy families \cite{cha2007,kullback2012}. However, these functions do not match perceptual dissimilarity well since they only compare weights of corresponding fixed bins \cite{rubner2000}. The families do not use the similarity information across neighboring bins such as adaptive neurons of weighted SOM patterns. When considering the information across bins, the Quadratic Form Distance (QFD) \cite{niblack1993} and the Earth Mover's Distance (EMD) \cite{rubner2000} are the most used functions. However, the QFD tends to overestimate the dissimilarity of patterns as the weight of each bin is simultaneously compared with weights across all bins \cite{rubner2000}. On the other hand, the EMD uses the ground distance of feature vectors across bins for the minimum weight flow providing better perceptual matches.

The EMD, as the other functions, aims to provide a numeric value to define only a notion of the overall resemblance of patterns. The overall dissimilarity itself cannot differentiate between weighted SOM patterns when the feature vector distance and the weight distribution have the same relation but different feature values such as $S_c$ and $S_d$ from $S_b$ in Fig. \ref{fig1}. This explains that such patterns can be further differentiated by the information of the pattern property in relation to the weight distribution. Moreover, it is difficult to identify the two different properties in $S_f$ that gain the change to $S_g$ (Fig. \ref{fig1}). In an attempt to handle these issues, we propose a visualization approach based on a metric using the EMD to measure the property change of weighted SOM patterns.

\section{Our Approach} \label{sec3}

In this section, we propose a visualization approach that uses a metric for measuring the property difference of weighted SOM patterns in order to capture the change information based on the property dissimilarity.  

\subsection{A Metric for Pattern Property Difference} \label{sec3.1}

In order to measure the pattern property difference, we introduce an extended function of the EMD, called Property EMD (PEMD). The PEMD measures the individual feature difference in the pattern change based on the capability of the EMD for the pattern dissimilarity measure. 

According to \cite{rubner2000}, the EMD between two weighted SOM patterns $P$ and $Q$ is defined as follows: 
\begin{equation}
EMD(P,Q) = \frac{\sum_{i=1}^n{\sum_{j=1}^n{{f_{ij}}{d_{ij}}}}}{\sum_{i=1}^n{\sum_{j=1}^n{f_{ij}}}},
\label{eq1}
\end{equation}
\noindent
where $d_{ij}$ is the ground distance function and $f_{ij}$ is the minimum cost flow under constraints: $\forall{i, j}$: $f_{ij} \geq 0$, $\forall{i}$: $\sum_{j=1}^n{f_{ij}} \leq {w_p}_i$, $\forall{j}$: $\sum_{i=1}^n{f_{ij}} \leq {w_q}_j$, and $\sum_{i=1}^n{\sum_{j=1}^n{{f_{ij}}}} =$ min$\{\sum_{i=1}^n{{w_p}_i}, \sum_{j=1}^n{{w_q}_j}\}$. The weighted SOM patterns $P$ and $Q$ are based on the same SOM; thus, they have the same number ($n$) of neurons and their weights equally sum to $1$. 

Based on the EMD, the difference of a feature $c_k$ in $Q$ for given $P$ can be measured by a function as follows:
\begin{equation}
PEMD_{c_k}(Q|P) = \frac{\sum_{i=1}^n{\sum_{j=1}^n{{f_{ij}}{d_{ij}}{({{c_k}_j}-{{c_k}_i})}}}}{\sum_{i=1}^n{\sum_{j=1}^n{f_{ij}}{d_{ij}}}}.
\label{eq2}
\end{equation}
\noindent
The direction of the feature change is accounted for by the difference measure. This distance is then defined as the resulting feature difference in the change normalized by the total work flow of the EMD. The feature difference is normalized to avoid favoring larger differences between pattern changes in the comparison. 

\begin{table}[b]
\centering
\caption{The scaled EMD and the property difference of the patterns in Fig. \ref{fig1}.}
\label{tbl2}
    \begin{tabular}{ | l | l || l | l | l |} \hline
    \textbf{Pattern Change} & \textbf{Scaled EMD} & \textbf{PEMD$_{R}$} & \textbf{PEMD$_{G}$} & \textbf{PEMD$_{B}$} \\ \hline
    \textbf{from $S_b$ to $S_c$} & 0.4491 & 0 & -1 & +1 \\ \hline
    \textbf{from $S_b$ to $S_d$} & 0.4491 & +1 & -1 & 0 \\ \hline
		\textbf{from $S_b$ to $S_e$} & 0.2245 & +0.5 & 0 & +0.5 \\ \hhline {=|=|=|=|=}
		\textbf{from $S_f$ to $S_g$} & 0.4082 & 0 & 0 & 0 \\ \hline
    \end{tabular}
\end{table}

The individual feature differences of the weighted SOM patterns in Fig. \ref{fig1} are measured by the PEMD for the property comparison. The EMD is also measured and scaled by the maximum EMD of the data space for the dissimilarity comparison. As the results show in Table \ref{tbl2}, the individual feature differences can be used to explain the property difference between the patterns $S_c$ and $S_d$ changed from $S_b$, which show the same EMD. However, it does not explain that the patterns $S_f$ and $S_g$ are not the same as shown by the EMD. This shows that the patterns can be further explained by possible feature values that gain the pattern change. Furthermore, it is difficult to compare the overall property dissimilarity if the dimensionality is high. Therefore, we propose a visualization approach in the next section for better analyzing the pattern change of high dimensional data based on the property difference by the PEMD.

\subsection{Visualization of Pattern Property Changes} \label{sec3.2}

\begin{figure}[tb]
\begin{center}
\includegraphics[width=\linewidth]{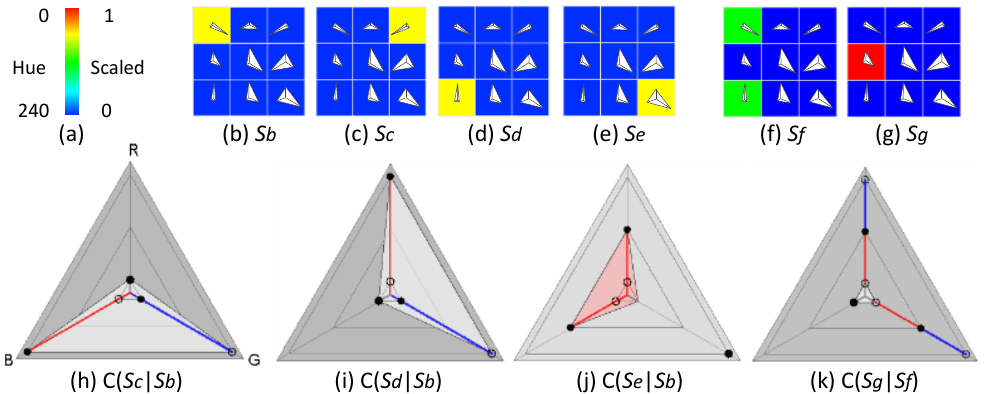}
\caption{$(a)$: The scaled hue color range. $(b)$ - $(g)$: Our views of the patterns in Fig. \ref{fig1}. $(h)$ - $(k)$: The visualized property changes of the patterns in Fig. \ref{fig1}. \label{fig:myapp}}
\end{center}
\end{figure}

Our visualization integrates colors and graphical objects, which are perceptually orthogonal \cite{wong1997}, to represent the pattern dissimilarity of high dimensional data. Hue colors and star glyph shape objects are used to view a weighted SOM pattern. The scaled hue colors in Fig. \ref{fig:myapp}$(a)$ are used to indicate high weight by red and low weight by blue. The graphical object mapping of prototype feature vector into a star glyph shape represents a neuron property in a SOM. A star glyph \cite{ward2008} has $m$ evenly angled branches emanating from a central point in the same ordering of $m$ dimensions. The length of each branch marks the value along the dimension it represents, and the value points are connected creating a bounded polygon shape. The patterns in Fig. \ref{fig1} are illustrated in Fig. \ref{fig:myapp}$(b)$ - $(g)$. As the shape is used as a single visual parameter, it is easier to recognize the property difference of neurons by only considering the shape variations in the fixed orientation \cite{ward2008}.  The perceptual dissimilarity of hue colors indicates a clear boundary of the weights. Thus, it facilitates the user selection of regions of interest for understanding the main properties of the weighted SOM patterns. 

The property dissimilarity between weighted SOM patterns can also be visualized by integrating colors and star glyph shape objects. As shown in Fig. \ref{fig:myapp}, a star glyph shape created by individual feature differences is imposed on the $3$-branch star glyph shape frame indicating the overall property dissimilarity. The PEMD value for each feature is scaled in the range of [$0.1, 0.9$] for the visualization. The average of the PEMD values is used to indicate the direction of the overall property change by applying it to the color saturation. The property shape is filled with the direction color; red for increase, blue for decrease and white for no change. The property direction color can differentiate between property changes if they have the same property shape but opposite directions of any individual feature changes. The possible value changes in each feature can be visualized depending on the user selection of regions (e.g. highly weighted regions). The possible value changes are indicated by red and blue for increase and decrease respectively (empty dots for the reference pattern and full dots for the changed pattern). The line with red or blue from the center to 0.1 indicates the direction of the individual feature change for increase or decrease, respectively. The overall property dissimilarity of the patterns can then be captured by the property shape variation with the change information in the fixed orientation. The EMD is also visualized by filling its gray scale in the frame for the dissimilarity comparison. Fig. \ref{fig:myapp}$(k)$ shows that $S_f$ and $S_g$ are not the same pattern by the EMD while there is no property difference in the change. This can be explained by observing the change ($C(S_g|S_f)$) which shows the possible changes in the features $R$ and $G$ by the same size increase and decrease while making no change to the pattern property. Fig. \ref{fig:myapp}$(h)$ and $(i)$ show that $S_c$ and $S_d$ obtained from $S_b$ are very different patterns by their different property shapes although they show the same EMD as the size and the direction color of the shapes.

In summary, our approach can measure the dissimilarity of weighted SOM patterns in terms of the pattern property. It measures the individual feature differences by the PEMD and captures the overall property dissimilarity by the visualization in the pattern change. It facilitates a simultaneous comparison of weighted SOM patterns and provides the information of how the patterns change.

\section{Experimental Results} \label{sec4}

In this section, we test our approach by applying it to the ecological domain data\footnote{The ecological features: $B1$ (Shredders), $B2$ (Filtering-Collectors), $B3$ (Collector-Gathers), $B4$ (Scrapers) and $B5$ (Predators) for Biological data set; $P1$ (Elevation), $P2$ (Slope), $P3$ (Stream Order), $P4$ (Embeddedness) and $P5$ (Water Temperature) for Physical data set. The feature values are all standardized for the total $130$ data.} \cite{giddings2009} for analyzing changes of output pattern by changes of input in the causal analysis. The physical and biological SOMs were trained using $10 \times 12$ hexagonal grids by the minimum values of quantization and topological errors. The physical input SOM was associated with the biological output SOM. Among the physical features, it is known that \textit{Embeddedness} ($P4$) has a strong impact on the biotic integrity \cite{novotny2005}. Thus, for our experiments, we varied the value of $P4$ in the physical input to examine the changes of the biological output. The value of $P4$ was increased by $1$ and $2$ standard deviations (SD) for the first and the second change, respectively, while the others were fixed at the initial value ($-0.5$SD). The standardized Z-score values of data used in the data analysis were converted to T-score values for visualization using our approach.

\begin{figure}[tb]
\begin{center}
\includegraphics[width=0.98\linewidth]{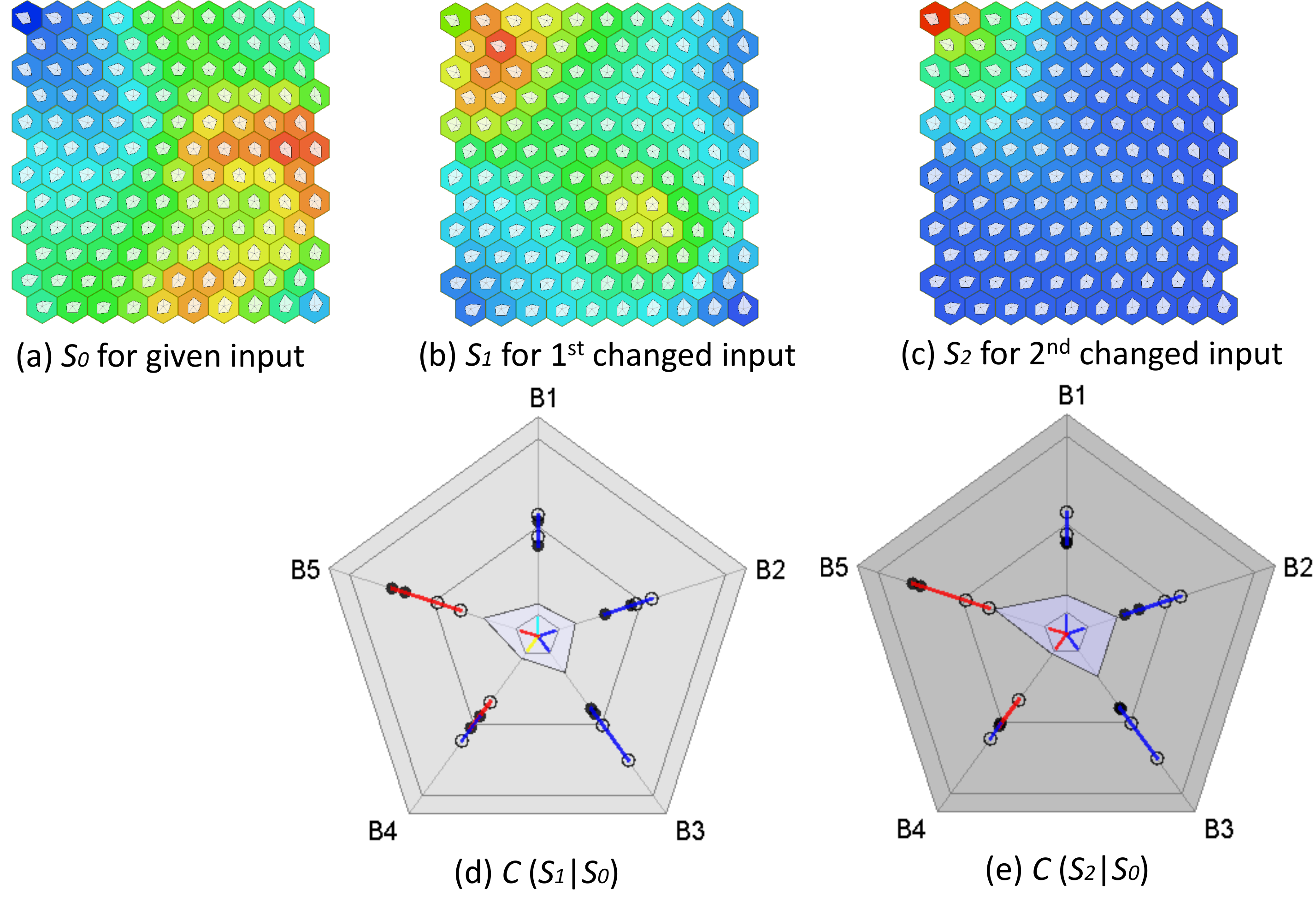} 
\caption{Our visualization approach. $(a)$ - $(c)$: The views of the weighted biological output SOM patterns. $(d)$ \& $(e)$: The pattern change views by the pattern differences. \label{fig:bioapp}}
\end{center}
\end{figure}

Fig. \ref{fig:bioapp} shows the weighted biological output SOM patterns for the physical input given in $(a)$, the first changed in $(b)$ and the second changed in $(c)$. More than one region is highly weighted in $S_0$ showing the high possibility of having different biological outputs for the given physical input. The reddish regions of each pattern are selected for analyzing the change. The difference of the patterns are measured and visualized using our approach in $(d)$ for the first change and $(e)$ for the second change. More information can be added in the view and we have added the significance information, measured on the difference of every weighted feature distribution using the Kolmogorov-Smirnov test \cite{press1992}. The insignificant changes are indicated by yellow and cyan while the significant changes are indicated by red and blue for increase and decrease, respectively. 

In Fig. \ref{fig:bioapp}$(d)$ and $(e)$, the user can observe that the second change $C(S_2 | S_0)$ is larger than the first change $C(S_1 | S_0)$ with a similar tendency of the property change. This can be explained by comparing the size, the direction color and the similarity of the property shape as well as the EMD gray scale in the frame. The individual change of each feature is also captured by the change information in the center of each branch. It shows that the changes in $B1$ and $B4$ become significant and the changes in $B1$, $B2$, $B3$ and $B5$ become larger when the value of $P4$ is increased. Based on the selected regions, the possible changes are also detailed on each branch. In particular, both increase and decrease are seen in $B4$, which cannot be provided by a quantitative measure. Throughout the experiments, the user can derive the impact of $P4$ (\textit{Embeddedness}) as its increase lowers the balance of the biological composition of the ecosystem. The causal effects are more effectively analyzed by considering all possible changes for well-informed decision making. Our approach supports this by detecting regions of interest and providing the change information visually; thus, it can be very useful for comparing the pattern changes in the process of causal analysis.

\section{Conclusion} \label{sec5}

In this paper, we have presented our approach for analyzing weighted output SOM pattern changes by input changes in causal analysis. We elucidated the idea of analyzing the change of weighted SOM patterns by comparing the dissimilarity of the pattern properties corresponding to the weight distributions. Our approach measures the property difference using a metric and uses a visualization to measure the property change of weighted SOM patterns. Throughout the experiments, we have shown that our approach is useful for measuring and comparing the pattern property in the change of weighted SOM patterns. We also facilitated exploring regions of user interest and capturing all possible changes to the pattern property. The experimental results show that our approach provides the property change information in an interactive and effective visual way when analyzing causal effects.

\end{document}